%% file: main.tex
\def\year{2023}\relax
\documentclass[letterpaper]{article} %
\usepackage{aaai23}  %
\usepackage{times}  %
\usepackage{helvet}  %
\usepackage{courier}  %
\usepackage[hyphens]{url}  %
\usepackage{graphicx} %
\usepackage{natbib}  %
\usepackage[labelfont=bf]{caption} %
\DeclareCaptionStyle{ruled}{labelfont=bf,labelsep=colon,strut=off} %

\usepackage[pagebackref=true,breaklinks=true,letterpaper=true,colorlinks,
    linkcolor=blue,
    filecolor=blue,
    citecolor = blue,bookmarks=false]{hyperref}

\usepackage{algorithm}
\usepackage{algpseudocode}
\usepackage{newfloat}
\usepackage{listings}

\usepackage{amssymb}
\usepackage{amsmath}
\usepackage{booktabs}
\usepackage{mathtools}
\usepackage{bbm}
\usepackage[normalem]{ulem}
\usepackage{comment}
\usepackage[capitalise]{cleveref}
\usepackage{color}
\usepackage{xspace,xfrac}
\usepackage{caption,xcolor}
\usepackage{multirow}
\usepackage{tabularx}
\usepackage{subfig}

\floatstyle{ruled}
\newfloat{listing}{tb}{lst}{}
\floatname{listing}{Listing}

\newcommand{\app}{\raise.17ex\hbox{$\scriptstyle\sim$}}

\title{Point-Teaching: Weakly Semi-Supervised Object Detection with
\\
with Point Annotations
}
\author {
Yongtao Ge$^1$, ~ Qiang Zhou$^2$, ~ Xinlong Wang$^1$, ~ Zhibin Wang$^2$, ~ Hao Li$^2$, ~ Chunhua Shen$^3$
\\[0.2cm]
\large
$^1$The University of Adelaide  ~~~~ $^2$Alibaba Damo Academy  ~~~~ $^3$Zhejiang University\thanks{
YG and QZ contributed equally. Part of this work was done when YG was visiting Alibaba and CS
was with The University of Adelaide. CS is the corresponding author.
}
}

\affiliations{
}

\def\eg{{\it e.g.}\xspace}
\def\ie{{\it i.e.}\xspace}

\def\MILImage{{image-wise MIL}}
\def\MilInst{{point-wise MIL}} %
\def\Ours{{Point-Teaching}\xspace}

\newcommand{\myparagraph}[1]{{\vspace{0.6em} \noindent
            \bf #1}}

\begin{document}

\maketitle
\input{1_Abstract}
\input{2_Introduction}

\input{3_RelatedWorks}
\input{4_Method}

\input{5_Experiments}
\input{6_Conclusion}
\input{7_Appendix}

\newpage
{\small
\bibliography{aaai23}
}
\end{document}

%% file: 1_Abstract.tex
 \begin{abstract}
    Point annotations are considerably more time-efficient than bounding box annotations. 
    However, how to use cheap point annotations to boost the performance of semi-supervised object detection
    remains largely unsolved. 
    In this work, we present \textbf{\Ours}, a weakly semi-supervised object detection framework to fully exploit the point annotations (WSSOD-P).
    Specifically, we propose a Hungarian-based point matching method to generate pseudo labels for point annotated images.
    We further propose multiple instance learning (MIL) approaches at the level of images and points to supervise the object detector with point annotations.
    Finally, we propose a simple-yet-effective data augmentation, termed point-guided copy-paste, to reduce the impact of the unmatched points.
    Experiments demonstrate the effectiveness of our method on %
    a few datasets and various data regimes.
    In particular, Point-Teaching outperforms the previous best method Group R-CNN by $3.1$ AP with $5\%$ fully label and $2.3$ AP with $30\%$ fully label on MS COCO dataset.
    We believe that our proposed framework can largely lower the bar of learning accurate object detector and pave the way for its broader applications. 

\end{abstract}

%% file: 2_Introduction.tex
\section{Introduction}

\iftrue
Great progress has been achieved in object detection and segmentation in recent years~\cite{FasterRCNN,YOLO,Focalloss,FCOS,maskrcnn,condinst,wang2021SOLO}.
Accurate object detectors can be trained using large fully-labeled datasets~\cite{MSCOCO,lvis}.
However, annotating large-scale object detection datasets are extremely expensive and time-consuming, as it requires the annotators to find all the objects of interest in the images and to draw a tight bounding box/segmentation mask for each of them.

How to train object detectors with fewer annotations has attracted increasing attention.
Weakly supervised object detection (WSOD) methods~\cite{Song2014,Gokberk2014,WSDDN,ContextLocNe,OICR,Ren2020} reduce the cost via replacing the box annotations with cheaper annotations, \textit{e.g.}, image-level categories, point clicks and squiggles.
 Semi-supervised object detection (SSOD) methods~\cite{CSD,STAC,UnbiasedTeacher,InstantTeaching,ISMT} train object detectors with a small amount of fully-labeled images and large-scale unlabeled images.
However, although both ways can reduce the annotation cost, the performance of the trained detectors is still far behind the fully-supervised counterpart.

\begin{figure}[!t]
    \centering
	    \includegraphics[width=0.86\linewidth]{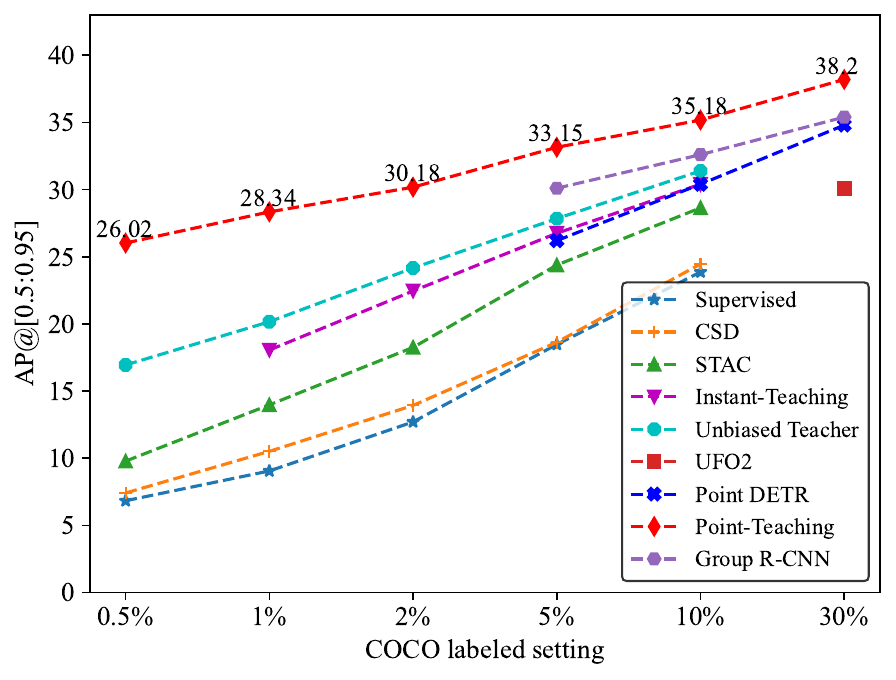}
        \caption{
        The performance ($\text{AP}$) of object detection on the COCO val. set under different labeled settings. The proposed \Ours{} is remarkably superior to other state-of-the-art SSOD and WSSOD-P methods.
        }
        \label{fig:comapre_sota}

\end{figure}

In this paper, we aim to train object detectors with considerably fewer annotations while achieving comparable performance with the fully-supervised counterpart.
To achieve this goal, there are two key problems: 1) what annotation formats to use and 2) how to train object detectors with such annotations.
A cheap but effective annotation format for object detection should be 1) simple to annotate, 2) convenient to store and use, 3) localization-aware.
Among various weak formats, point click annotation stands out as it meets all the requirements.
Point click provides a stronger prior of object location compared with image-level category annotation. Meanwhile, it does not require detailed and expensive location information such as object bounding box or segmentation masks, thus being considerably more time-efficient.
According to \cite{Papadopoulos-iccv} and \cite{cheng2021pointly}, a box annotation takes 7 seconds while a point annotation takes 0.8-0.9 seconds.
To achieve the  best balance of detection performance and annotation cost,
we adopt mixed annotation formats to construct the training dataset.
In the following, we use \textit{point annotated setting} to represent such a dataset which comprises a small number of fully annotated images and massive point annotated images.
Under this setting, we are able to obtain abundant annotations in a relatively cheaper manner~\cite{Crowdsourcing_anno,Human-machine-collaboration}.

To fully utilize both the limited box annotations and abundant point annotations,
we propose a novel weakly semi-supervised object detection framework, termed \textbf{\Ours}.
Inspired by Mean Teacher~\cite{mean-teacher} and Unbiased Teacher~\cite{UnbiasedTeacher}, we construct a Student model and a Teacher model with the same architecture.
In each training iteration, weakly augmented point-labeled images are fed to the Teacher model to generate reliable pseudo bounding boxes.
The Student is then optimized on fully labeled and pseudo labeled images with strong augmentation.
The Teacher is updated via Exponential Moving Average (EMA) of the Student.
Within this basic framework, we propose three key components tailored for point annotations.
First, we propose the {hungarian-based point matching} method to generate pseudo labels for point annotated images.
A spatial cost and a classification cost are introduced to find the bipartite matching between point annotations and predicted box proposals.

We further propose multiple instance learning (MIL) approaches at the level of images and points to supervise the object detector with point annotations.
Inspired by previous WSOD works \cite{WSDDN,ContextLocNe,OICR,Ren2020}, we perform {image-wise MIL} via treating the whole image as a bag of object proposals. These proposals are aggregated for predicting all presented classes in the image, supervised by image-level labels during the training.
To leverage the location information of point annotations,
we propose {point-wise MIL}, which selects the highest detection score proposal with the same class label as the only positive and suppresses the rest proposals as negatives around the given point annotation.
Finally, we propose the {point-guided copy-paste} augmentation strategy.
The motivation is that there still exist some point annotations that have not been matched any proposals after the point matching.
To further utilize those unmatched points, we maintain an online object bank and paste same-class objects to unmatched point annotations during the training. The point-guided copy-paste makes the distribution of generated pseudo labels closer to that of the ground truth.

\fi

Experiments demonstrate the effectiveness of our method on different datasets and various data regimes. Point-Teaching has the following advantages:
1) Our method can boost the performance of existing SSOD methods, \eg, over the strong semi-supervised baseline method Unbiased Teacher~\cite{UnbiasedTeacher},
our detector achieves significant improvements of $9.1$ AP with $0.5\%$ fully labeled data on MS COCO.
2) Our method outperforms all existing WSSOD-P methods in all data regimes by a large margin.
In particular, when using $30\%$ fully labeled data from MS COCO, our method outperforms previous state-of-the-art WSSOD-P method Group R-CNN~\cite{zhang2022group} by $2.3$ AP and Point DETR~\cite{point-as-query} by $3.4$ AP.

Our main contributions are summarized as follows:
\begin{itemize}
    \itemsep 0.2cm
\item
We propose a simple and effective training framework for weakly semi-supervised object detection, termed { \bf \Ours{}}, which integrates point annotations into semi-supervised learning. The key components of \Ours{} include {Hungarian-based point-matching} approach, {image-wise and instance-wise MIL} loss, and {point-guided copy-paste} augmentation.

\item Extensive experiments are conducted on MS-COCO and VOC datasets to verify the effectiveness of our method. {\Ours} significantly outperforms the existing methods \cite{point-as-query,zhang2022group} and greatly narrows the gap between weakly semi-supervised and fully-supervised object detectors.

\item We further extend \Ours{} from WSSOD-P to weakly semi-supervised instance segmentation (WSSIS) and weakly-supervised instance segmentation (WSIS), setting a strong baseline for the two challenging tasks.

\end{itemize}

%% file: 3_RelatedWorks.tex
\begin{figure*}[t]
    \centering
    \small
    \includegraphics[width=1.0\linewidth]{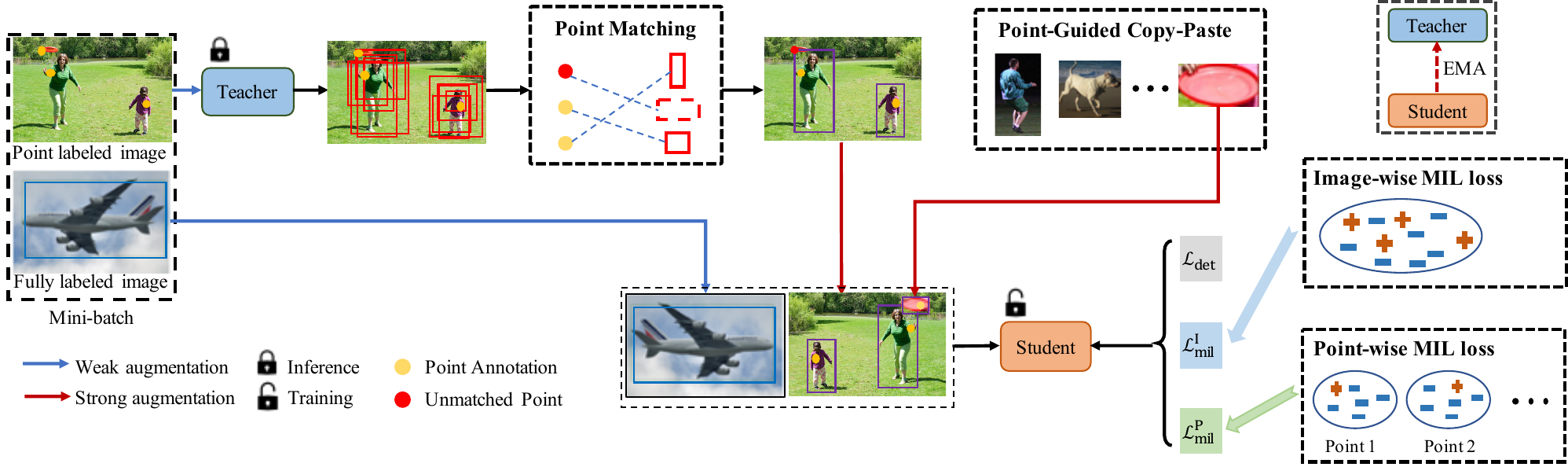}
    \caption{The training process of \Ours{}. In each training iteration, the Teacher model first generates pseudo box annotations for the point-labeled images with weak augmentation. The Student model is then trained on fully-labeled images with weak augmentation and point-labeled images with strong augmentation.
    The Teacher model is gradually updated by the student model via EMA.
    Image-wise MIL loss constructs a bag containing all predicted boxes, and the number of positive boxes in the bag is uncertain.
    The point-wise MIL loss constructs a bag for each annotated point,
    and there is only one positive box in these bags.
    }
    \label{fig:training}
\end{figure*}

\section{Related Work}

\myparagraph{Fully-supervised object detection.}
With the large-scale fully annotated detection datasets~\cite{MSCOCO,lvis}, existing modern detectors have obtained great improvements in the object detection task.
These detectors can be divided into three categories: two-stage detectors~\cite{FasterRCNN,DoubleHead}, one-stage detectors~\cite{YOLO,SSD,FCOS} and the recently end-to-end detectors~\cite{DETR,DeformableDETR,PSS}.
Faster RCNN is a popular two-stage detector that first generates region proposals and then refines these proposals in the second stage.
Unlike two-stage detectors, one-stage detectors, such as YOLO~\cite{YOLO} and FCOS~\cite{FCOS}, directly output dense predictions of classification and regression without refinement.
Recently, DETR~\cite{DETR} introduces the transformer encoder-decoder architecture to object detection and effectively removes the need for many hand-craft components, e.g. predefined anchors and non maximum suppression (NMS) post-processing.
Despite the great success, these detectors are trained with large amounts of expensive fully-labeled data. Therefore, a lot of work has been proposed to reduce the annotation cost.

\myparagraph{Weakly-supervised object detection.}
There exist many WSOD works that focus on training object detector with weakly-labeled data.
Most previous studies have two phases: proposal mining and proposal refinement.
The proposal mining phase is formulated as the MIL problem to implicitly mine object locations with image-level labels. The proposal refinement phase aims at refine the object location with the predictions from the proposal mining phase.
WSDDN~\cite{WSDDN} proposes a two-stream network to simultaneously perform region selection and classification. The region level scores from these
two streams are then element-wise multiplied and transformed to image-level scores by summing over all regions.
Following WSDDN~\cite{WSDDN}, ContextLocNe~\cite{ContextLocNe} introduces context information.
OICR~\cite{OICR} presents a multi-stage refinement strategy to avoid the MIL detector be trapped into the local minimum.
PCL~\cite{PCL} proposes to refine instance classifiers by clustering region proposals in an image to different clusters.
MIST~\cite{Ren2020} proposals a multiple instance self-training framework.
OIM~\cite{OIM} effectively mining all possible instances by introducing information propagation on spatial and appearance graphs.
However, propagating image-level weak supervision to instance-level training data inevitably involves a large amount of noisy information and the performance of these methods are limited.

\myparagraph{Semi-supervised object detection.}
Besides WSOD, SSOD addresses the problem by using large amount of unlabeled data, together with a small set of labeled data.
One popular SSOD technique is consistency regularization, which aims to regularize the detector's prediction with an image of different augmentations. CSD~\cite{CSD} enforces the detector to make consistent predictions on an input image and its horizontally flipped counterpart. ISD~\cite{ISD} proposes an interpolation-based method for SSOD.
Another emerging SSOD approach is pseudo labeling, where a teacher model is trained on labeled data to generate pseudo labels on unlabeled data, and a student model is then trained on both labeled and pseudo labeled data. STAC~\cite{STAC} pre-trains a model on labeled data and fine-tunes it on both labeled and unlabeled data iteratively.
Instance-Teaching~\cite{InstantTeaching} introduces a co-rectify scheme for alleviating confirmation bias of pseudo labels.
Unbiased Teacher~\cite{UnbiasedTeacher} proposes a class-balance loss to address the class imbalance issue in pseudo-labels and refine the teacher model via Exponential Moving Average (EMA).

\myparagraph{Weakly semi-supervised object detection.}
Image-level annotation is a kind of weak annotation compared to box annotation. However, it is not optimal for object detection since the lack of instance-level information.
Recently, point supervision~\cite{UFO2,point-as-query,Papadopoulos-cvpr,Papadopoulos-iccv} has been employed in WSSOD.
Papadopoulos \text{et al.}~\cite{Papadopoulos-cvpr,Papadopoulos-iccv} collect click annotation for the PASCAL VOC dataset and train an object detector through iterative multiple instance learning.
UFO$^2$~\cite{UFO2} proposes a unified object detection framework that can handle different forms of supervision simultaneously, including box annotation and point annotation. Point DETR~\cite{point-as-query} extends DETR~\cite{DETR} by adding a point encoder and thus can convert point annotations to pseudo box annotations.
Group R-CNN~\cite{zhang2022group} proposes to use instance-level proposal grouping for each point annotation and thus can get a high recall rate.
In this paper, we follow this setting and introduce several methods for improving the performance of WSSOD-P.

%% file: 4_Method.tex
\section{Method}

\subsection{Preliminaries}

\myparagraph{Problem definition.}
In this work, we study weakly semi-supervised object detection
under the \textit{point annotated setting}, in which the dataset consists of a small set of fully annotated images $D_F=\{(I_{i}, \hat{b}_i)\}_{i=1}^{N_F}$ and a large set of point annotated images $D_P=\{(I_i, \hat{p}_i)\}_{i=1}^{N_P}$.
$N_F$ and $N_P$ are the number of fully labeled and point labeled images respectively. $I_i$ denotes fully or point labeled images.
For fully annotated images, the annotation $\hat{b}_i$ includes box coordinates $(\hat{b}_i^{x_1}, \hat{b}_i^{y_1}, \hat{b}_i^{x_2}, \hat{b}_i^{y_2})$ and class label $\hat{b}_i^{l}$.
For point annotated images, the annotation $\hat{p}_i$ includes point location $(\hat{p}_i^{x}, \hat{p}_i^{y})$ and class label $\hat{p}_i^{l}$.
For point annotated images, we only need to randomly annotate one point for each object instance, thereby the annotation cost can be greatly reduced.

\subsection{Overall Architecture}
For a fair comparison, we take Faster RCNN with FPN~\cite{FasterRCNN} and ResNet-50 backbone~\cite{ResNet} as our baseline object detector.
Compared to the original Faster RCNN network~\cite{FasterRCNN}, we add two additional parallel branches to the RCNN head, termed Objectness-I branch and Objectness-P branch,  respectively.
The Objectness-I branch is used to suppress the likelihood of inconsistent classification predictions with image-level annotations, and is optimised with \MILImage{} loss.
The Objectness-P branch is developed to measure the quality of pseudo boxes at point level, and is supervised with \MilInst{} loss.
The key difference is that Objectness-I branch select the most probable region proposals for each class from the image-level bag with different classes. While Objectness-P branch performs binary classification to select the most probable region proposal from the point-level bag that only contains region proposals of the same class.

The training pipeline of \Ours{} is represented in~\cref{fig:training}. Inspired by Mean Teacher~\cite{mean-teacher} and Unbiased Teacher~\cite{UnbiasedTeacher}, there are two models with the same architecture, a {student} model and a {teacher} model.
In each training iteration, weakly augmented point-labeled images from the dataset $D_P$ are firstly fed to the {Teacher} for reliable pseudo labels; the {Student} is then optimized by labels from fully-labeled dataset $D_F$ and pseudo labels generated from the {Teacher} with strong augmentation; Finally, the {Teacher} is updated by EMA of the {Student}. Different from the original Unbiased Teacher~\cite{UnbiasedTeacher}, there are three key components within the proposed framework: \textit{hungarian-based point matching strategy, point supervision with image-wise and instance-wise MIL loss, and point-guided copy-paste augmentation.}

\subsection{Point Matching}
\label{sec:point_matching}

%
%

%
%
%
%
%
%

In order to find the best matching between the annotated points and the predicted boxes, \ie, to choose the best box prediction for each point annotation,
we propose a simple point matching method, termed {hungarian-based point matching}.
Specifically, we design two types of matching costs between annotated points and predicted boxes: a spatial cost and a classification cost.
For the spatial cost, we consider two factors: 1) Predicted boxes that share the same class label with the given point annotation should have a low cost. 2) Predicted boxes with point annotations inside lead to a low cost.
For the classification cost, higher confidence scores of the Classification branch and Objectness-P branch lead to a lower cost.

Formally, the cost matrix $\mathcal{L}_{\text{match}} \in \mathbb{R}^{N_p \times N_b}$ is defined as:

\begin{equation}
\begin{aligned}
    \mathcal{L}_{\text{match}}(i, j) &=
    \underbrace{ (1 - \mathbbm{1} [\hat{p}_i~\text{in}~b_j] \cdot \mathbbm{1} [\hat{p}_i^l = b_j^l]) }_{\text{spatial}~\text{cost}} \\
    &+  \underbrace{ (1 - \sigma(\boldsymbol{s}_{j,\hat{p}_i^l}) \cdot \sigma^{\text{P}}{(\boldsymbol{s}^{\text{P}}_{j,1}))}}_{\text{classification}~\text{cost}},
\end{aligned}
\label{eq:cost}
\end{equation}
where $i$ is the index of the annotated points, and $j$ is the index of the predicted boxes.
$\mathcal{L}_{\text{match}}(i, j)$ denotes the matching cost between the annotated point $\hat{p}_i$ and the predicted box $b_j$.
$N_p$ is the number of annotated points, and $N_b$ is the number of predicted boxes.
$\hat{p}_i^l$
indicates the class label of the annotated point $\hat{p}_i$, $b_j^l$ indicates the class label of the predicted box $b_j$.
We use $\boldsymbol{s} \in \mathbb{R}^{N_b \times (C+1)}$ %
and $\boldsymbol{s}^{\text{P}}\in \mathbb{R}^{N_b \times 2}$ to denote the outputs of Classification and Objectness-P branches, respectively, where $C$ denotes the number of categories excluding background.
$\sigma (\cdot)$ represents the \textit{softmax} operation on the Classification output along the second dimension.
$\sigma^{\text{P}} (\cdot)$ represents the \textit{softmax} operation on the Objectness-P output along the second dimension.

Once the cost matrix is defined, the {point matching} problem could be mathematically formulated as a bipartite matching problem as:
\begin{equation}
\hat{\pi}=\underset{\pi \in \mathfrak{S}_{N_b}}{\arg \min } \sum_{i}^{N_p} \mathcal{L}_{\operatorname{match}}\left(i, \pi(i)\right),
\label{eq:matching}
\end{equation}
where $\pi \in \mathfrak{S}_{N_b}$ indicates a permutation of $N_b$ elements. This optimal assignment can be solved with the Hungarian algorithm~\cite{kuhn1955hungarian}.

\subsection{MIL Loss for Images and Points}
In this section, we present the overall loss function $\mathcal{L}$ of \Ours{} framework.

\begin{equation}
\mathcal{L}=\mathcal{L}_{\text {det }}+\lambda_{1} \mathcal{L}_{\text {mil}}^{\text {I}}+\lambda_{2} \mathcal{L}_{\text {mil}}^{\text {P}}.
\label{eq:total_loss}
\end{equation}
As shown in~\cref{eq:total_loss}, the overall loss $\mathcal{L}$ consists of three parts: $\mathcal{L}_{\text {det }}$, $\mathcal{L}_{\text {mil}}^{\text {I}}$ and $\mathcal{L}_{\text {mil}}^{\text {P}}$, respectively.
$\mathcal{L}_{\text {det }}$ represents the losses of the original object detector, \eg classification loss and regression loss in RPN and ROI head of Faster RCNN.
$\mathcal{L}_{\text {mil}}^{\text {I}}$ is \MILImage{} loss, which is proposed in WSDDN~\cite{WSDDN}.
$\mathcal{L}_{\text {mil}}^{\text {P}}$ is our proposed  \MilInst{} loss, which is defined below.
$\lambda_1$ and $\lambda_2$ are hyper-parameters used to balance these three loss terms.

\myparagraph{Image-wise MIL loss.}
\label{sec:mil_image}
Given the point annotations, we can easily obtain image-level labels $\{\hat{\phi}_c, ~c=1, \cdots, C\}$.
Image labels can help improve the performance of object detection in two ways.
First, for categories that do not present in the image, the image-level supervision could help decrease the confidence score of the corresponding predicted boxes.
Second, it helps detect the objects of the categories that present in the image.

Taking hundreds of predicted boxes as a bag, we only know the class labels of the entire bag and do not know the individual class label of each predicted box.
Let us denote by $\boldsymbol{s}, \boldsymbol{s}^{\text{I}} \in \mathbb{R}^{N_b \times C}$ the output of Classification branch and Objectness-I branch, respectively;
$\sigma^{\text{I}} (\cdot)$ the \textit{softmax} operation on the first dimension.
We share ROI features of box proposals with two fully-connected layers and then produce two score matrices $\sigma\left(\boldsymbol{s}\right), \sigma^{\mathrm{I}}\left(\boldsymbol{s}^{\mathrm{I}}\right) \in \mathbb{R}^{N_b \times C}$ by Classification branch and {Objectness-I branch}, respectively.
Then the element-wise product of the two score matrix is a new score matrix $X^{\mathrm{s}} \in \mathbb{R}^{N_b \times C}$, which can be formulated as: $X^{\mathrm{s}} = \sigma\left(\boldsymbol{s}\right) \odot \sigma^{\mathrm{I}}\left(\boldsymbol{s}^{\mathrm{I}}\right)$.
Finally, a sum pooling is applied to obtain image-level classification scores:
\begin{equation}
\begin{split}
\mathbf{\phi}_{c} &= \sum_{i=1}^{N_b} X_{i c}^{\mathrm{s}}
                  = \sum_{i=1}^{N_b} \left[\sigma\left(\boldsymbol{s}_{i, c}\right) \odot \sigma^{\mathrm{I}}\left(\boldsymbol{s}_{i, c}^{\mathrm{I}}\right)\right].
\end{split}
 \label{eq:image-score}
\end{equation}

Based on the obtained image-level labels and image-level classification scores, the introduced image-wise MIL loss is defined as the sum of binary cross-entropy loss across all categories:
\begin{equation}
    \mathcal{L}_{\text {mil}}^{\text {I}} = -\sum_{c=1}^{C} \left ( \hat{\phi_c} ~\text{log}(\phi_c) + (1 - \hat{\phi_c})~ \text{log}(1 - \phi_c) \right ) \text{,}
 \label{eq:image-mil-loss}
\end{equation}
where $C$ is the number of categories, $\hat{\phi_c} \in\{0,1\}^{C}$ is the image-level one-hot labels, and $\phi_c$ denotes the predicted image-wise classification scores.

\myparagraph{Point-wise MIL loss.}
\label{sec:mil_inst}
To perform multiple instance learning at point level,
we construct a bag with part of the predicted boxes for each annotated point, as shown in \cref{fig:training}.
For example, the constructed bag $\Psi_i$ for the annotated point $\hat{p}_i$ consists of those predicted boxes that enclose point $\hat{p}_i$ and have the same class label as $\hat{p}_i$.
In other words, $\Psi_i = \{b_j ~|~ \mathbbm{1}[\hat{p}_i ~\text{in} ~b_j] \cdot \mathbbm{1}[\hat{p}_i^l = b_j^l] \}$, in which $\hat{p}_i^l$ denotes the class label of annotated point $\hat{p}_i$, and $b_j^l$ denotes the class label of the predicted box $b_j$.
Unlike the bag of image-wise MIL loss, there is only one positive box proposal inside $\Psi_i$, defined as the best predicted box corresponding to the annotated point $\hat{p}_i$.
Assuming we know how to calculate the bag-level confidence score $\varphi_i$ for bag $\Psi_i$, we can define the proposed \MilInst{} loss as:
\begin{equation}
    \mathcal{L}_{\text {mil}}^{\text {P}} = -\sum_{i=1}^{N_p}
    {\rm log}(\varphi_i),
    \label{eq:mil_inst}
\end{equation}
where $N_p$ denotes the number of annotated points, and $\mathcal{L}_{\text {mil}}^{\text {P}}$ is the sum of the binary cross-entropy loss for all annotated points.

Next, we explain how to compute the bag-level confidence score $\varphi_i$ corresponding to bag $\Psi_i$.
To help find out the best box proposal inside $\Psi_i$, we add the Objectness-P branch.
This branch performs binary classification to predict whether the box is the best prediction inside bag $\Psi_i$, and its output is denoted as $\mathbf{s}^{\text{P}} \in \mathbb{R}^{N\times2}$.
Since there should be only one positive box inside the bag $\Psi_i$, we use a slightly different way to compute $\varphi_i$. As shown in \cref{eq:mil_inst_score}:
\begin{equation}
    \varphi_i = \sum_{k=1}^{|\Psi_i|}\left [ \sigma(\boldsymbol{s}_{k,\hat{p}_i^l}) \odot \sigma^{\text{P}}{(\boldsymbol{s}^{\text{P}}_{k,1})} \odot \prod_{m!=k}\sigma^{\text{P}}{(\boldsymbol{s}^{\text{P}}_{m,0})}\right ],
    \label{eq:mil_inst_score}
\end{equation}
in which $\sigma (\cdot)$ and $\sigma^{\text{P}} (\cdot)$ denote \textit{softmax} operation as described earlier, ${|\Psi_i|}$ indicates the number of predicted boxes in bag $\Psi_i$.
Comparing \cref{eq:image-score} and \cref{eq:mil_inst_score}, we can find that the element-wise multiplication before accumulation is different.
Taking the $k^{\text{th}}$ predicted box in bag $\Psi_i$ as an example.
In addition to multiplying the positive confidence score of the two branches (i.e., $\sigma(\boldsymbol{s}_{k,\hat{p}_i^l}) \cdot \sigma^{\text{P}}{(\boldsymbol{s}^{\text{P}}_{k,1})}$), we also multiply the negative confidence scores of the Objectness-P
branch of the remaining boxes in bag $\Psi_i$ (i.e., $\prod_{m!=k}\sigma^{\text{P}}{(\boldsymbol{s}^{\text{P}}_{m,0})}$). With the help of negative confidence scores, the proposed point-wise MIL loss can encourage that each bags have and only have one positive box with the highest positive confidence score, while the positive confidence score of remaining boxes is suppressed.
The pseudo-code of \MilInst{} loss based on PyTorch is provided in the supplementary.

\subsection{Point-Guided Copy-Paste}
\label{sec:copy-paste}
During the point matching, we observe that \app6\% of the annotated points are not matched with any predicted boxes, and these unmatched points usually correspond to difficult instances to be detected (e.g. instances from minority classes).
Ignoring these unmatched points may cause the class imbalance of the generated pseudo boxes.
The confirmation bias in pseudo boxes further reinforces the imbalance issue.
To alleviate the impact of these unmatched points, we propose a simple data augmentation strategy termed \textit{point-guided copy-paste}.
Different from naively copying ground-truth boxes from one labeled image to another unlabeled image like Simple Copy-Paste~\cite{copy-paste},
we maintain a dynamic object bank as depicted in \cref{fig:training}, which will be updated with ground truth object patches (cropped based on box annotation) from fully labeled images and pseudo object patches from point labeled images during each training iteration.
For each unmatched point after the point matching stage, we randomly select an object patch with the same class label from the object bank, and paste the selected patch near the point on the original image. The effectiveness of point-guided copy-paste augmentation is verified in~\cref{sec:ablations}.

%% file: 5_Experiments.tex
\section{Experiment}

\setlength{\tabcolsep}{4pt}
\begin{table*}[t]
\begin{center}
\resizebox{\linewidth}{!}{
\begin{tabular}{@{}r|c|c|c|c|c|c|c@{}}
\toprule
Method           & Type & 0.5\%  & 1\%  & 2\%  & 5\%  & 10\% & 30\% \\ \midrule
Supervised   & FSOD   & 6.83 $\pm$ 0.15      & 9.05 $\pm$ 0.16     & 12.70 $\pm$ 0.15    & 18.47 $\pm$ 0.22    & 23.86 $\pm$ 0.81 & 31.99 $\pm$ 0.82 \\ \midrule
CSD~\cite{CSD}    & SSOD  & 7.41 $\pm$ 0.21     & 10.51 $\pm$ 0.06    & 13.93 $\pm$ 0.12    & 18.63 $\pm$ 0.07   & 24.46 $\pm$ 0.08 & - \\ 
STAC~\cite{STAC} & SSOD  & 9.78 $\pm$ 0.53      & 13.97 $\pm$ 0.35    & 18.25 $\pm$ 0.25   & 24.38 $\pm$ 0.12   & 28.64 $\pm$ 0.21 & - \\ 
Instant-Teaching~\cite{InstantTeaching}  & SSOD  & -         & 18.05 $\pm$ 0.15    & 22.45 $\pm$ 0.15    & 26.75 $\pm$ 0.05    & 30.40 $\pm$ 0.05 & -   \\ 
Unbiased Teacher~\cite{UnbiasedTeacher}  & SSOD  & 16.94 $\pm$ 0.23     & 20.16 $\pm$ 0.12    & 24.16 $\pm$ 0.07   & 27.84 $\pm$ 0.11    & 31.39 $\pm$ 0.10  & -   \\ \midrule

Point DETR~\cite{point-as-query}  & WSSOD-P   & -     & -    & -   & 26.2    & 30.4  &  34.8 \\
Group R-CNN~\cite{zhang2022group}  & WSSOD-P   & -     & -    & -   & 30.1    & 32.6  &  35.4 \\
\Ours{} & WSSOD-P   & \textbf{26.02} $\pm$ \textbf{0.09} & \textbf{28.34} $\pm$ \textbf{0.02} &  \textbf{30.18}  $\pm$ \textbf{0.08}  &  \textbf{33.15} $\pm$ \textbf{0.07} &  \textbf{35.18} $\pm$ \textbf{0.09} & \textbf{38.20} $\pm$ \textbf{0.10} \\ \bottomrule

\end{tabular}
}
\end{center}
\caption{Comparison of our proposed \Ours{} with other SSOD (without point-level labels) and WSSOD-P (with point-level labels) methods on COCO val. set. All these models use R50-FPN as the backbone network. \Ours{} are trained with a batch size of 64 (32 fully-labeled images and 32 point-labeled images) and $180$k iterations. Note that the upper bound of $100\%$ fully supervised model is {$40.2$ AP~\cite{detectron2}}.
}
\label{tbl:main_with_semi}
\end{table*}
\setlength{\tabcolsep}{1.4pt}

\subsection{Datasets}
We mainly benchmark our proposed method on the large-scale dataset
MS-COCO~\cite{MSCOCO}. 
Following~\cite{point-as-query}, we synthesize the point annotations by
randomly sampling a point inside the annotated box.
Then we discard the box annotations of point-labeled images.
Specifically, We randomly selected 0.5\%, 1\%, 2\%, 5\% , 10\% and 30\% from the $118$k labeled images as the fully-labeled
set, and the remainder is used as the point-labeled set. Model performance is evaluated on the COCO2017 val set. We also conduct experiments on PASCAL VOC~\cite{everingham2010pascal}. The VOC results are in Appendix.

\subsection{Implementation Details}

We implement our proposed Point-Teaching framework based on the Detectron2 toolbox~\cite{detectron2}. 
For fair comparison with existing works~\cite{STAC,InstantTeaching,UnbiasedTeacher}, we take Faster RCNN with FPN~\cite{FasterRCNN} as our object detector and ResNet-50~\cite{ResNet} as backbone.
The feature weights are initialized by the ImageNet pretrained model. 
Our method mainly contains three hyperparameters: $\tau$, $\lambda_1$ and $\lambda_2$, which indicates the score threshold of the pseudo boxes, the loss weight of image-wise MIL loss and the loss weight of point-wise MIL loss, respectively.
We set $\tau=0.05$, $\lambda_1 = 1.0$ and $\lambda_2 = 0.05$ unless otherwise specified. 

We use $AP_{50:95}$ (denoted as AP) as evaluation metric. On Pascal VOC, the models are trained for $40$k iterations on $8$ GPUs and with batch size $32$, which contains $16$ box-labeled images and $16$ point-labeled images respectively.
Other training and testing details are same as the original Unbiased-Teacher~\cite{UnbiasedTeacher}.

\subsection{Ablation Study}\label{sec:ablations}

When conducting ablation experiments, we choose 1\% MS-COCO protocol and take a quick learning schedule of $90$k iterations and a smaller batch size of 32, containing 16 box-labeled images and 16 point-labeled images, respectively.

\begin{table}[!ht]
\centering
\small
\begin{tabular}{@{}r|c|c@{}}
\toprule
Point Location & $\text{AP}_{50:95}$   & $\text{AP}_{50}$ \\ \midrule
random      & 25.18 & 48.26   \\ 
center      & 25.19 & 48.28 \\ \bottomrule
\end{tabular}
\caption{Comparison of the effectiveness of the point location on the COCO val. set. `random' and `center' indicate the annotation location on objects.}
\label{tbl:point_location}
\end{table}

\myparagraph{Effects of point location.}
We verify the effectiveness of point annotation location to Point-Teaching between two point location schemes: center point and arbitrary point on objects. 
As shown in~\cref{tbl:point_location}, when using center point on objects as our annotation, Point-Teaching achives 25.2 AP.
While we randomly sample point inside the box annotation, the performance only slightly drops $0.01\%$ AP, showing that Point-Teaching is insensitive to the location of point annotation.

\begin{table}[!ht]
\centering
\small    
\begin{tabular}{@{}r|c|c@{}}
\toprule
Point Matching & $\text{AP}_{50:95}$   & $\text{AP}_{50}$ \\ \midrule
None           & 20.2 & 36.5 \\ 
Hungarian      &  \textbf{25.2}    &   \textbf{48.3}   \\ \bottomrule
\end{tabular}
\caption{Comparison of detection accuracy on the COCO val. set by varying the point matching methods when selecting pseudo box annotations.}
\label{tbl:point_matching}
\end{table}  

\myparagraph{Effects of point matching.}
We explore the impact of our proposed Hungarian-based point matching method on the model performance. In this experiment, we set the loss weights of $\lambda_1$ and $\lambda_2$ to 0.
As shown in~\cref{tbl:point_matching}, when point matching is not used, the model reaches 20.2 AP, as reported in Unbiased-Teacher~\cite{UnbiasedTeacher}. Taking point annotations into consideration and using our proposed Hungarian matching, the model reaches 25.2 AP, which improves the AP with 5.0 absolute points.

\begin{table}[!h]
\centering
\small
\begin{tabular}{@{}c|c|c|c@{}}
\toprule
$\lambda_1$ & $\lambda_2$ & $\text{AP}_{50:95}$   & $\text{AP}_{50}$ \\ \midrule
0.5             &       &   25.00   &   47.88  \\ 
1.0    & 0      &   \textbf{25.66}   &    \textbf{49.04}  \\
1.5             &       & 25.01 & 48.46 \\  \bottomrule
\end{tabular}
\caption{Comparison of detection accuracy on COCO val. set when varying the loss weight $\lambda_1$ of \MILImage{} loss.}
\label{tbl:mil_image}
\end{table}

\myparagraph{Loss weight $\lambda_1$ of \MILImage{} loss.}
We conduct experiments to explore the effect of loss weight $\lambda_1$ of \MILImage{} loss. In these experiments, we use Hungarian-based point matching and set the loss weight $\lambda_2$ of \MilInst{} loss to 0, \ie the Objectness-P branch is not optimized during training and the Objectness-P score is removed in \cref{eq:cost} when computing the cost matrix. As shown in~\cref{tbl:mil_image}, when loss weight $\lambda_1$ reaches 1.0, the model achieves 
the highest AP.
If not specified, in other experiments, we will set $\lambda_1$ to 1.0 by default.

\begin{table}[!ht]
\centering
\small
\begin{tabular}{@{}c|c|c|c@{}}
\toprule
 $\lambda_1$ & $\lambda_2$    & $\text{AP}_{50:95}$  & $\text{AP}_{50}$ \\ \midrule
    & 0.025     & 25.85 & 49.63 \\ 
0   & 0.05      & \textbf{25.97} & \textbf{49.90} \\ 
    &  0.1      & 25.74 & 49.77 \\ 
    & 0.15      & 25.40 & 48.82 \\ \bottomrule
\end{tabular}
\caption{Comparison of detection accuracy on COCO val. set when varying the loss weight $\lambda_2$ of \MilInst{} loss.}
\label{tbl:mil_instance}
\end{table}

\myparagraph{Loss weight $\lambda_2$ of \MilInst{} loss.}
We conduct experiments to explore the effect of loss weight $\lambda_2$ of \MilInst{} loss. In these experiments, we use Hungarian-based point matching and set the loss weight $\lambda_1$ of \MILImage{} loss to 0. 
As shown in~\cref{tbl:mil_instance}, when loss weight $\lambda_2$ reaches 0.05, the model achieves the highest AP. If not specified, in other experiments, we will set $\lambda_2$ to 0.05 by default.

\label{sec:threshold}
\begin{table}[!ht]
\centering
\small
\begin{tabular}{@{}c|c|c|c|c@{}}
\toprule
$\tau$ & $\lambda_1$ & $\lambda_2$ & $\text{AP}_{50:95}$ & $\text{AP}_{50}$ \\ \midrule
0.01 &          &           &  26.24     & \textbf{50.71}  \\
0.05 &    1.0   &    0.05   &  \textbf{26.28}     & 50.44  \\
0.1 &           &           &  26.17     & 50.01   \\ 
0.15 &          &           &  26.19     & 49.94      \\ \bottomrule
\end{tabular}
\caption{Comparison of detection accuracy on the COCO val. set when varying the score threshold $\tau$}
\label{tbl:score_thr}
\end{table}
\myparagraph{Score threshold $\tau$.}
The score threshold $\tau$ is used to filer out low quality pseudo boxes. We conduct experiments to explore the effect of score threshold $\tau$.
When conducting these experiments, we use Hungarian-based point matching and set the loss weights of $\lambda_1$ and $\lambda_2$ to 1.0 and 0.05 respectively.
As shown in~\cref{tbl:score_thr}, when $\tau$ reaches 0.05, the model achieves the highest AP.
If not specified, in other experiments, we set $\tau$ to 0.05 by default.

\begin{table}[!h]
\centering
\small
\begin{tabular}{@{}c|c|c|c|c@{}}
\toprule
H. PM & I. MIL & P. MIL & P. CP & $\text{AP}_{50:95}$   \\ \midrule
                &               &               &               & 20.2 \\ 
$\checkmark$    &               &               &               & 25.2  \\ 
$\checkmark$    & $\checkmark$  &               &               & 25.7 \\ 
$\checkmark$    &               & $\checkmark$  &               & 26.0  \\ 
$\checkmark$    & $\checkmark$  & $\checkmark$  &               & 26.3  \\
$\checkmark$    & $\checkmark$  & $\checkmark$  & $\checkmark$  & \textbf{27.3}  \\ \bottomrule
\end{tabular}
\caption{The effect of each element proposed in this work. H. PM indicates hungarian-based point matching, I. MIL denotes image-wise MIL loss and P. MIL indicates point-wise MIL loss, P. CP indicates point-guided copy-paste augmentation.}
\label{tbl:element}
\end{table}

\myparagraph{Factor-by-factor experiment.}
We conduct a factor-by-factor experiment on our proposed Hungarian-based point matching, image-wise MIL loss, point-wise MIL loss and {point-guided copy-paste}.
As shown in~\cref{tbl:element}, each element of our proposed \Ours{} has a positive impact on the performance of the model. When all these elements are combined, the model reaches the highest performance, \ie, $27.3$ AP.

\subsection{Comparison with State-of-the-art Methods}

We verify our method with previous studies on \text{COCO-standard} dataset. As shown in~\cref{tbl:main_with_semi}, our method consistently surpass all previous SSOD models (CSD, Instance Teaching, Unbiased Teacher) and WSSOD-P models (Group R-CNN and Point DETR) in all data regimes that 0.5\% to 30\% data are fully-labeled. The results also indicate that Point-Teaching is robust to fewer fully-label data compared to previous methods, \eg Point-Teaching outpeforms Point DETR~\cite{point-as-query} by $6.95$ AP and Group RCNN~\cite{zhang2022group} by $3.1$ AP under 5\% COCO labeled data.

\subsection{Extensions: Point-Teaching for Instance Segmentation}\label{sec:wsis}
In order to demonstrate the generality of Point-Teaching, we extend our framework to weakly semi-supervised instance segmentation. In this experiment, Mask RCNN with ResNet-50 backbone is used as our detector and only fully-labeled data has box and mask annotations. As shown in~\cref{tbl:main_instance_segmtation}, Point-Teaching significantly improve the performance in all data regimes. This result indicates that Point-Teaching can benefit from only a small amount of mask annotations. Thus, it is a promising approach to reduce the annotation cost in weakly semi-supervised instance segmentation task.

\begin{table}[t]
\centering
\resizebox{0.68\linewidth}{!}{
\begin{tabular}{@{}r|c|c|c|c|c@{}}
\toprule
\multirow{2}{*}{Method} & \multirow{2}{*}{Backbone} & \multicolumn{4}{c}{COCO Labeled Setting} \\ \cmidrule(l){3-6} 
                        &                           & 1\%    & 2\%    & 5\%    & 10\%  \\ \midrule
Supervised ($\text{AP}^{\text{mask}}$)              & R50-FPN                   & 10.8  & 14.5  & 18.7  & 22.6 \\ \midrule
{\Ours} ($\text{AP}^{\text{mask}}$)                   & R50-FPN                   & 23.5  & 25.9  & 30.7 & 33.3 \\ \bottomrule
\end{tabular}
}
\caption{\Ours{} for weakly semi-supervised instance segmentation on COCO val. set. Results are reported with mask $\text{AP}_{50:95}$. All models are trained with a batch size of 32 (16 fully-labeled images and 16 point-labeled images) and $180$k iterations.}
 \label{tbl:main_instance_segmtation}
\end{table}

\begin{table}[!t]
\centering
\resizebox{0.70\linewidth}{!}{
\begin{tabular}{@{}r|c|c}
\toprule
\multirow{2}{*}{Method} & \multirow{2}{*}{Backbone}  & \multicolumn{1}{c}{COCO Labeled Setting} \\ \cmidrule(l){3-3} 
                                        &             & 30\%    \\ \midrule
Supervised ($\text{AP}^{\text{mask}}$)  & R50-FPN     &  22.1  \\ \midrule
{\Ours} ($\text{AP}^{\text{mask}}$)     & R50-FPN     &  28.0 (\textcolor{red}{$\uparrow$5.9})  \\ \bottomrule
\end{tabular}
}
\caption{\Ours{} for weakly-supervised instance segmentation on COCO val. set. Results are reported with mask $\text{AP}_{50:95}$.
}
\label{tbl:wsis}
\end{table}

We further extend our framework to weakly supervised instance segmentation. 
In this scenario, we supervise the instance segmentation training with \textbf{only box and point annotations}. Specifically, we train Mask RCNN with ResNet-50 under 30\% COCO labeled setting. The whole training pipeline contains two stage. In the first stage, we use proposed Point-Teaching framework to get a well-trained teacher model. In the second stage, we fix the teacher model with zero EMA update rate and use the proposed hungarian-based point matching method to generate pseudo bounding boxes, and the student model is supervised with both annotated and pseudo-annotated boxes with three additional loss terms, \eg point loss, project loss~\cite{hsu2019bbtp,tian2021boxinst} and pairwise loss~\cite{tian2021boxinst}. More details about loss functions can be found in the supplementary materials. As shown in~\cref{tbl:wsis}, Point-Teaching achieves 28.0 mask AP without mask annotation, outperforming the supervised baseline by 5.9 AP. 

%% file: 6_Conclusion.tex
\section{Conclusion}
In this work, we presented Point-Teaching, a novel weakly semi-supervised framework for object detection and instance segmentation. It can effectively leverage point annotation with the proposed {hungarian-based point matching} strategy, {image-wise MIL} loss, {point-wise MIL} loss, and {point-guided copy paste} augmentation. These contributions enable our framework significantly outperforms all previous works by a large margin in all data regime settings. We hope our work can inspire the community to design more practical object detectors with limited human annotations.

%% file: 7_Appendix.tex
\appendix

\section{Results on VOC dataset}

We conduct experiments on \text{VOC07\&12} dataset. As represented in Table~\ref{tbl:compare_sota_voc}, our methods achieves $83.03$ $\text{AP}_{50}$, $2.5$ $\text{AP}_{50}$ improvement against the Unbiased Teacher. Moreover, Point-Teaching is only $0.47$ $\text{AP}_{50}$ and $1.82$ AP below the fully-supervised model trained on \text{VOC07\&12} dataset, which demonstrates Point-Teaching has fully-supervised-level capacity.

\begin{table}[!ht]
\centering
\resizebox{0.98\linewidth}{!}{
\begin{tabular}{@{}r|c|c|c@{}}
\toprule
Method                                  & Labeled  & $\text{AP}_{50}$   &  $\text{AP}_{50:95}$     \\ \midrule
Supervised                              & VOC07       & 72.63 & 42.13 \\
Supervised                              & VOC07\&12  & 83.50 & 56.39    \\\midrule
CSD~\cite{CSD}                             & VOC07     & 74.70 & -        \\
STAC~\cite{STAC}                            & VOC07     & 77.45  & 44.64  \\
Instant-Teaching~\cite{InstantTeaching} & VOC07          & 79.20     & 50.00   \\
Unbiased Teacher~\cite{UnbiasedTeacher} & VOC07            & 80.51    & 54.48    \\ \midrule
Point-Teaching                          & VOC07     & \textbf{83.03} & \textbf{54.57}   \\ \bottomrule
\end{tabular}
}
\caption{Comparison of our proposed \Ours{} with other SSOD methods on Pascal VOC dataset. The unlabeled or point-labeled images are VOC12 \textit{trainval} set.
}
\label{tbl:compare_sota_voc}
\end{table}

\section{Percentage of Point Labels}
\begin{figure}[!h]
    \centering
    \includegraphics[width=0.45\textwidth]{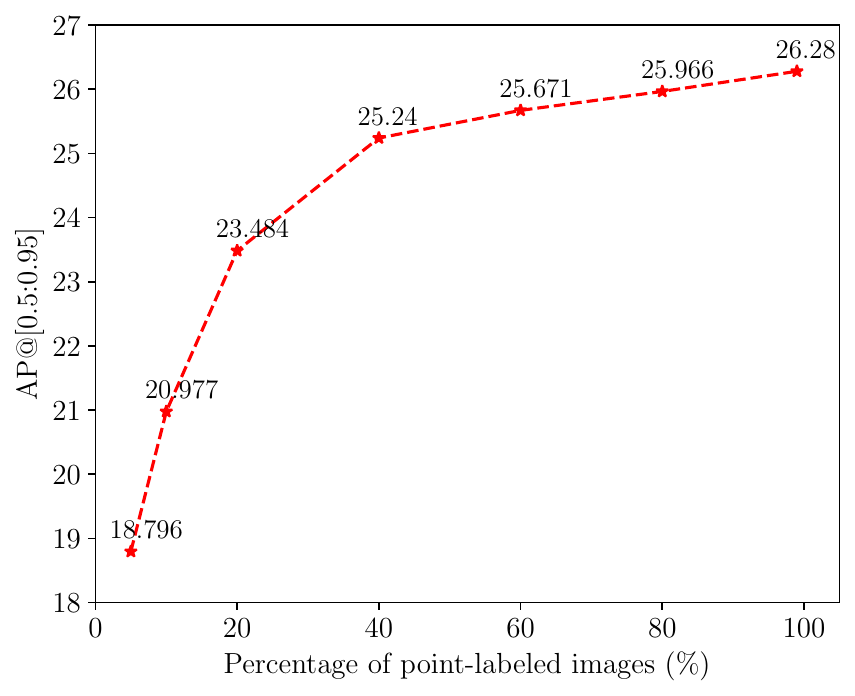}
    \caption{The curve of detection performance with relative to the percentage of the point-labeled images.}
    \label{fig:percentage}
\end{figure}

In this section, we study the effect of the percentage of point-labeled images. As shown in Table 1, %
keeping the percentage of fully-labeled images unchanged at 1\% COCO setting, when the percentage of point-labeled images is gradually increased, the detection performance of our Point-Teaching improves accordingly.
Even if the number of point-labeled images is 99 times of the fully-labeled images, the performance improvement of our Point-Teaching has no obvious saturation trend, as shown in Figure~\ref{fig:percentage}.

\begin{table*}[!h]
\centering
\resizebox{0.6572\linewidth}{!}{
\begin{tabular}{@{}r|c|c|c|c|c|c|c|c@{}}
\toprule
\multirow{2}{*}{Data setting} & fully-labeled & \multicolumn{7}{c}{point-labeled}        \\ \cmidrule(l){2-9}
                                  & 1\%           & 5\% & 10\% & 20\% & 40\% & 60\% & 80\% & 99\% \\ \midrule
AP   &  9.05 &  18.80   &  20.98    & 23.48     & 25.24      & 25.67     & 25.97   & 26.28   \\ \bottomrule
\end{tabular}
}
\label{tbl:unlabel_percent}
\caption{Comparison of detection performance on COCO val. set when changing the percentage of point-labeled images. In these experiments, we keep the percentage of fully-labeled images constant at 1\% COCO setting.}
\end{table*}

\section{Error Analysis}
As mentioned in the paper, previous SSOD methods need to apply confidence thresholding to filter low-confidence predicted boxes, \eg Unbiased Teacher select $\tau=0.7$ for thresholding.
However, it is not necessary for our method to carefully tune the confidence threshold $\tau$, and we set $\tau=0.05$ in all of our experiments.
To empirically examine the effectiveness of Point-Teaching on pseudo-labeling, we evaluate the \text{AP} of pseudo labels on MS-COCO val set.
According to our experiment, the AP of Point-Teaching's pseudo labels is $28.77$ when trained on 10\% MS-COCO protocol, which outperforms Unbiased Teachers~\cite{UnbiasedTeacher} by $3.0$ AP.

\begin{figure*}
\subfloat[
Unbiased Teacher
]{
    \begin{minipage}[t]{0.45\linewidth}
    {
    \begin{center}
    \includegraphics[width=0.5\linewidth]{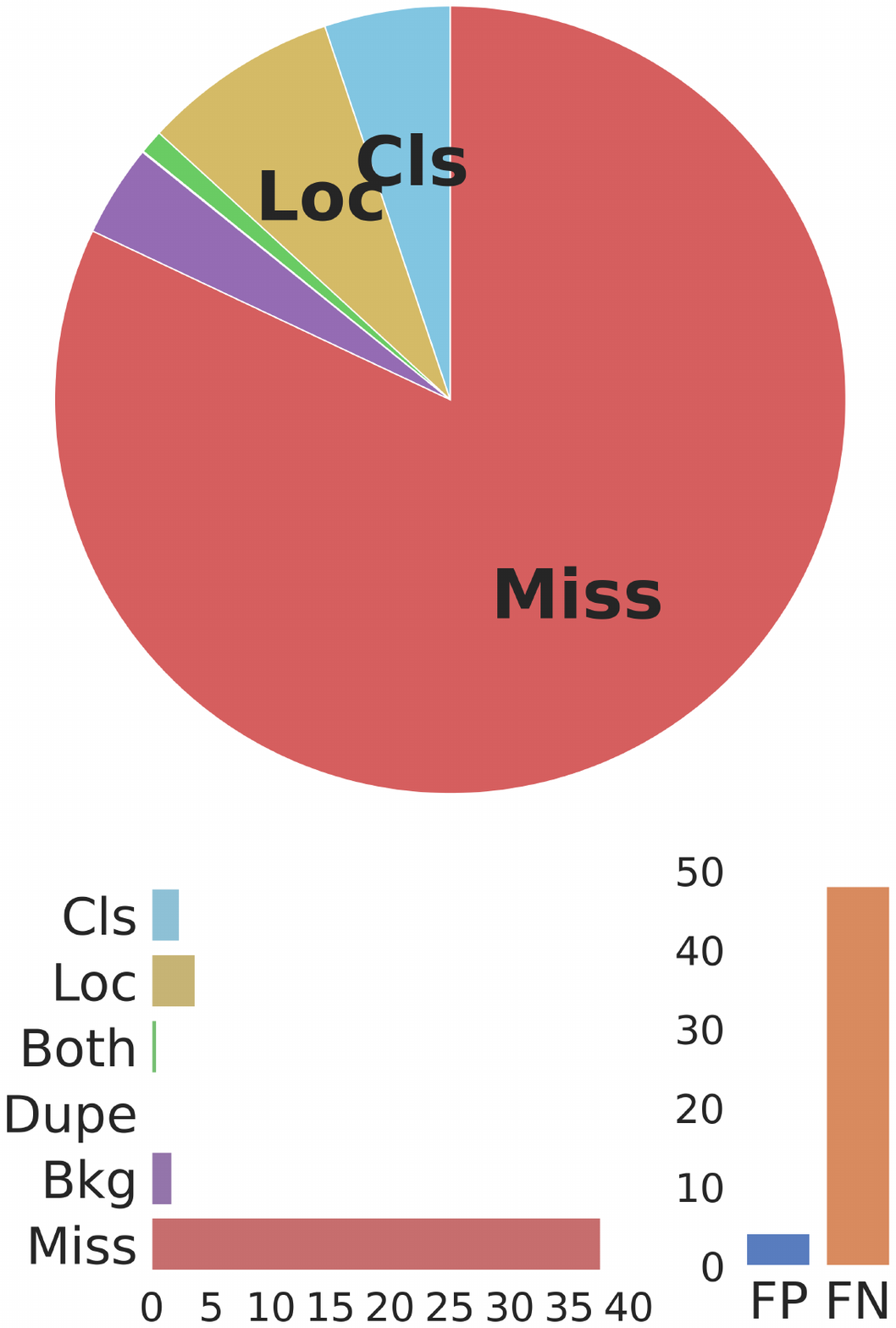}
    \end{center}
    }
    \end{minipage}
}
\subfloat[
Point-Teaching
]{
    \begin{minipage}[t]{0.45\linewidth}
    {
    \begin{center}
    \includegraphics[width=0.5\linewidth]{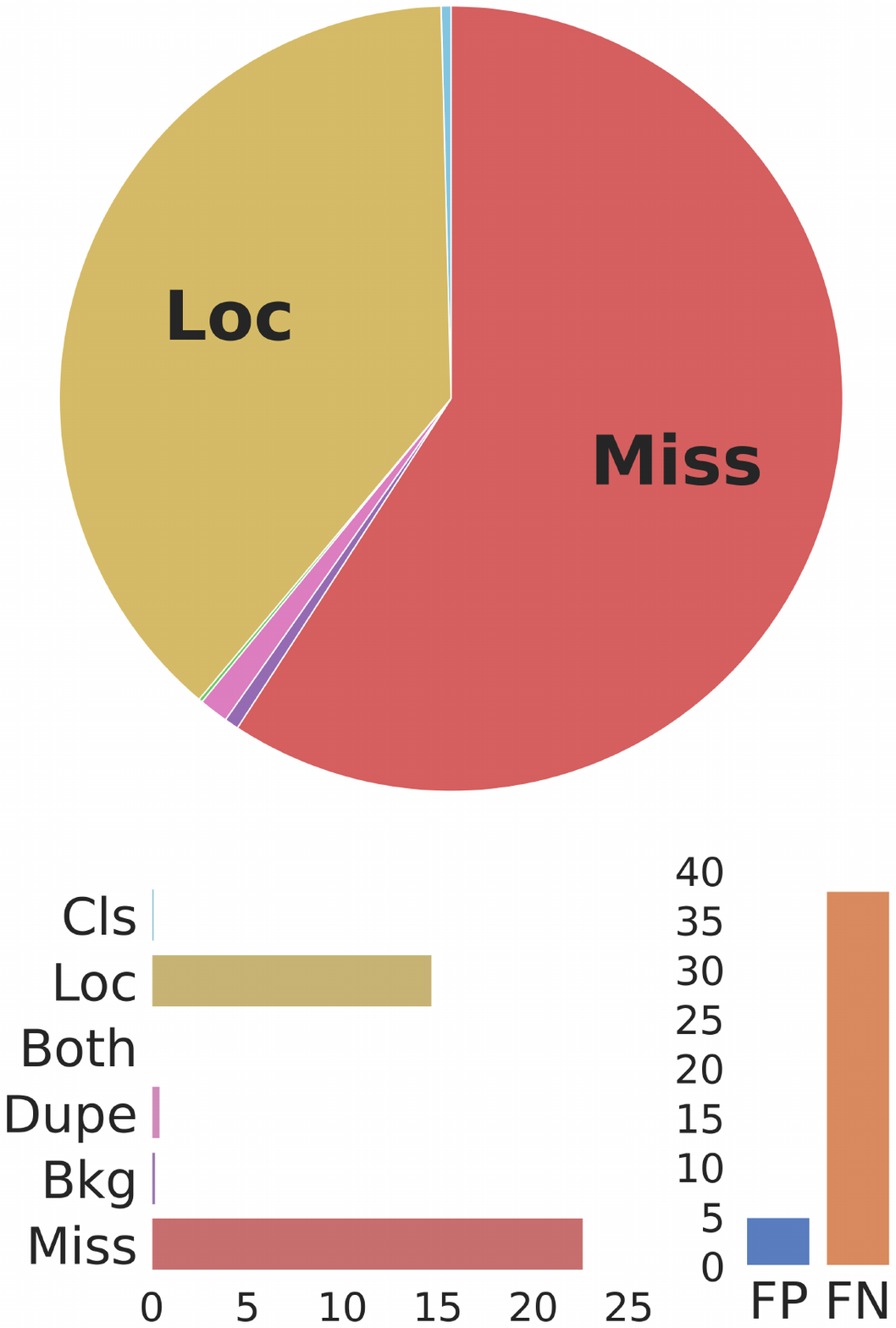}
    \end{center}
    }
    \end{minipage}
}
    \caption{Error analysis of generated pseudo-boxes by TIDE~\cite{tide-eccv2020}. Different error types: \textbf{Cls}: localized correctly but classified incorrectly, \textbf{Loc}: classified correctly but localized incorrectly, \textbf{Both}: both cls and loc error, \textbf{Dupe}: duplicate detection error, \textbf{Bkg}: detected background as foreground, \textbf{Miss}: missed ground truth error.}
    \label{fig:error}

\end{figure*}

We also analyze the errors of generated pseudo boxes by TIDE~\cite{tide-eccv2020} in~\cref{fig:error}. Compared with Unbiased Teacher~\cite{UnbiasedTeacher}, our method has lower error rates in all error types except for \textbf{Loc} error. This is caused by the use of lower confidence threshold $\tau$.

\section{Pseudo Code of Point-wise MIL Loss}
In this section, we provide pseudo-code of point-wise MIL loss based on PyTorch for better understanding. For more details, please refer to Algorithm~\ref{alg:mil_inst_code}.

\section{Weakly-Supervised Instance Segmentation}
As mentioned in the paper, \Ours can easily extend to weakly-supervised instance segmentation task with only point and box annotations. Inspired by Boxinst~\cite{tian2021boxinst}, we leverage both ground truth box and pseudo box annotations as supervision to train Mask RCNN. Specifically, the horizontal and vertical projections of the predicted mask is supervised by the projection loss term; the pairwise similarity of predicted masks are supervised by pairwise affinity loss term (pixels with similar colors are very likely to have the same label). We refer readers to~\cite{tian2021boxinst} for more details.

Besides the box supervision, we leverage the point annotations during the training. Given a box proposal, we first extract a feature map (ROI feature) by ROI Align~\cite{maskrcnn} from the backbone. Then, following~\cite{cheng2021pointly}, we sample prediction features in the locations of ground truth points annotations from the ROI feature on the grid using bilinear interpolation. Last, cross-entropy loss is applied as supervision for prediction features and point labels.

\begin{algorithm*}[!t]
\caption{Pseudo-code of point-wise MIL loss in a PyTorch-like style.}
\label{alg:mil_inst_code}
\definecolor{codeblue}{rgb}{0.25,0.5,0.5}
\definecolor{codegreen}{rgb}{0,0.5,0}
\definecolor{codeblue}{rgb}{0.25,0.5,0.5}
\definecolor{codegray}{rgb}{0.6,0.6,0.6}
\lstset{
  backgroundcolor=\color{white},
  basicstyle=\fontsize{7.5pt}{8.5pt}\fontfamily{lmtt}\selectfont,
  columns=fullflexible,
  breaklines=true,
  captionpos=b,
  commentstyle=\fontsize{8pt}{9pt}\color{codegray},
  keywordstyle=\fontsize{8pt}{9pt}\color{codegreen},
  stringstyle=\fontsize{8pt}{9pt}\color{codeblue},
  frame=tb,
  otherkeywords = {self},
}

\begin{lstlisting}[language=python]
import torch
from torch.nn import functional as F

def point_wise_loss(scores, scores_2, proposal_boxes, point_xy_list, point_label_list):
    """
    scores: output of 'Classification' branch.
            Tensor of shape [N, C+1], and forground starts from 0.
    scores_2: output of 'Objectness-P' branch.
            Tensor of shape [N, 2].
    proposal_boxes: proposals from RPN. Tensor of shape [N, 4].
    point_xy: coordinate of groundtruth point.
    point_label: label of groundtruth point.
    """
    scores = F.log_softmax(scores, dim=1)
    scores_2 = F.log_softmax(scores_2, dim=1)
    mil_loss_points = []

    for point_xy, point_label in zip(point_xy_list, point_label_list):
        # extract proposals with point inside
        idxs_p = ((point_xy[0] >= proposal_boxes[:, 0])
                    & (point_xy[0] <= proposal_boxes[:, 2])
                    & (point_xy[1] >= proposal_boxes[:, 1])
                    & (point_xy[1] <= proposal_boxes[:, 3])).nonzero().reshape(-1)

        scores_p = scores[idxs_p]
        scores_2_p = scores_2[idxs_p]
        log_fg_prob = scores_p[:, point_label].detach() + scores_2_p[:, 1]
        log_bg_prob = scores_2_p[:, 0]

        eye = torch.eye(len(log_fg_prob), dtype=torch.float32,
                                          device=log_fg_prob.device)
        log_prob = (eye * log_fg_prob[None, :] + \
                    (1 - eye) * log_bg_prob[None, :]).sum(dim=-1)
        max_ = log_prob.max()
        point_bag_log_prob = torch.log(torch.exp(log_prob - max_).sum()) + max_

        mil_loss_points.append(-point_bag_log_prob)

    mil_loss_points = torch.stack(mil_loss_points).mean()
    return mil_loss_points

\end{lstlisting}
\end{algorithm*}